\documentclass[11pt]{article}

\usepackage[utf8]{inputenc}
\usepackage[T1]{fontenc}
\usepackage{geometry}
\usepackage{setspace}
\usepackage{graphicx}
\usepackage{amsmath,amssymb}
\usepackage{booktabs}
\usepackage{hyperref}
\hypersetup{colorlinks=true, linkcolor=blue, citecolor=blue, urlcolor=blue}
\usepackage{xr}
\usepackage{natbib}
\setcitestyle{numbers,square}
\usepackage{caption}
\usepackage{subcaption}
\usepackage{xcolor}
\usepackage{enumitem}
\usepackage{listings}
\title{\textbf{Agentic Calibration of Grey-Box Simulation Models: \\
An LLM-Driven Alternative \\
}}

\author{%
  David Gómez-Guillén\textsuperscript{1,2}\thanks{Corresponding author: \href{mailto:dgomez@iconcologia.net}{dgomez@iconcologia.net}} \and
  Mireia Díaz\textsuperscript{1,3} \and
  Josep Lluis Arcos\textsuperscript{4} \and
  Jesús Cerquides\textsuperscript{4} 
}

\date{%
  \textsuperscript{1}Unit of Infections and Cancer -- Information and Interventions (UNIC-I\&I), Cancer Epidemiology Research Program, Catalan Institute of Oncology-IDIBELL, L'Hospitalet de Llobregat, Spain \\
  \textsuperscript{2}Autonomous University of Barcelona (UAB), Spain \\
  \textsuperscript{3}Centro de Investigación Biomédica en Red de Epidemiología y Salud Pública (CIBERESP), Ministerio de Ciencia, Innovación y Universidades, Madrid, Spain \\
  \textsuperscript{4}Artificial Intelligence Research Institute, IIIA-CSIC, Spain \\
}

\begin{document}

\maketitle
\thispagestyle{empty}

% =============================================================
\begin{abstract}
\noindent\textbf{Background.}
Calibration of simulation models used in cost-effectiveness analysis
(CEA) is a constrained optimization problem in which model
evaluations are expensive, the parameter space can be
high-dimensional, and the search must respect clinical plausibility
constraints. Although the simulation
code is fully available to the analyst, making the model grey-box rather than strictly black-box, the joint effect of multiple
parameters on the outputs remains difficult to predict analytically.
Classical derivative-free optimizers such as Nelder--Mead (NM) are
simple to deploy but sample-inefficient, particularly under
constraints. Modern Bayesian Optimization (BO) methods such as
BAxUS achieve competitive solutions with far fewer evaluations but
require non-trivial modeling machinery for constraint handling.

\medskip\noindent\textbf{Methods.}
We introduce an agentic calibration method in which a large
language model (LLM) acts as the optimizer. At each
iteration, the LLM is provided with the history of
parameter guesses, the simulated outputs, the calibration targets,
and a structured breakdown of per-group residuals, and proposes
the next parameter vector together with a rationale.
Constraints are incorporated as
a plain-language section of the system prompt, requiring no additional surrogate models or constraint-specific machinery. The resulting
optimization trace is human-readable and
can be audited after the fact, in contrast to the opaque
internal mechanisms of GP surrogates or simplex-based methods. We evaluate the agentic method, NM, and
BAxUS on a 10-dimensional anal cancer simulation model under both
unconstrained and clinically constrained calibration.

\medskip\noindent\textbf{Results.}
Under unconstrained calibration, the agentic method achieves substantially lower
median best error than BAxUS and NM, while requiring 16 model
evaluations, compared to 110 for BAxUS and several hundred to
several thousand for NM. Under constrained calibration, the agentic method reaches comparable error levels and both outperform NM, with the
agentic method requiring no additional implementation to support constraints. These results are obtained at the cost of increased inference time per iteration. 

\medskip\noindent\textbf{Conclusions.}
Agentic calibration achieves competitive performance with substantially
fewer model evaluations, and constraint
handling is essentially free at the modeller-facing interface through
simple textual specifications rather than additional modelling
machinery. The main trade-off lies in increased per-iteration
inference cost, balanced by a reduced number of simulations, making
the approach particularly suitable when simulation time dominates, as
is often the case in CEA. Beyond performance, the per-iteration
rationale makes the search auditable, so its decisions can be
scrutinised and justified, an attractive property for regulated
health-technology-assessment settings.

\end{abstract}

\newpage

% =============================================================
\section{Introduction}

Health-economic decision modelling, and in particular
cost-effectiveness analysis (CEA), relies on simulation-based models
whose parameters are only partially identified from observational
data. Calibrating such models, that is, finding parameter vectors that
reproduce a set of epidemiological or clinical targets,
is a frequent bottleneck in cost-effectiveness
analyses \citep{stout2009,vanni2011}. From an optimization
standpoint, calibration is challenging for three reasons: each evaluation
requires running the full simulation, the objective surface is often non-convex, and clinical knowledge typically
imposes constraints that the optimizer must respect.

An important characteristic of CEA calibration is that the
simulation code is fully available to the analyst. Unlike the
strictly black-box setting commonly assumed in the optimization
literature \citep{shahriari2016,garnett2023,papenmeier2022}, a CEA
calibration is more accurately described as a grey-box problem. In
this setting, the structure of the model (parameter meanings, transition
probabilities, target metrics) is known and interpretable, yet the
joint effect of multiple parameters on the simulated outputs
remains difficult to predict analytically. The available knowledge
about the model can be substantial: for many calibrations, a
domain expert can reason qualitatively about the direction in
which a parameter should be adjusted given a residual pattern. Such
problem-specific reasoning is precisely the kind of information
that black-box optimizers cannot exploit.

Classical derivative-free algorithms, such as Nelder--Mead (NM)
\citep{neldermead1965} and simulated annealing,
are well understood and easy to deploy, but their sample efficiency
is poor and they have no principled way of incorporating soft
inequality constraints beyond rejection sampling or penalty terms.
Bayesian Optimization \citep{shahriari2016,garnett2023} replaces
direct search with a probabilistic surrogate, typically a Gaussian
Process (GP), and an acquisition function balancing exploration
and exploitation. Modern high-dimensional BO variants such as
BAxUS \citep{papenmeier2022} and SAASBO \citep{eriksson2021}, together with
constraint-aware extensions \citep{gardner2014}, currently
represent the state of the art for expensive black-box
optimization in moderate-to-high dimensions.

A third option has recently become viable. Large language models
(LLMs) trained on text and code perform competitively as
in-context optimizers
\citep{yang2024large,liu2024llmevopro,llm-calibration-hydrology} in applications ranging from
prompt optimization to materials discovery
\citep{ramos2024review,song2024position}. Existing work in this
line has primarily targeted strictly black-box settings. The
present paper instead applies an LLM-driven optimizer to a
grey-box calibration problem, where the code, the parameter
semantics and the clinical interpretation of the targets are all
made available to the model in natural language. Three features make
this approach particularly relevant in the calibration setting. First, the
LLM can ingest the full natural-language description of a problem,
including the meaning of each parameter, the targets, and the constraints,
and use this information to reason about which residuals matter
most, in a way analogous to an experienced human modeller. Second,
the analyst-facing interface remains compact: constraints, target
adjustments, and reparameterisations are all changes to the
prompt rather than to the surrogate or acquisition code. Third, the
method is auditable: because the LLM states a written rationale for
each proposed parameter vector, the calibration produces a
human-readable trace of why the search moved as it did. This contrasts
with the opaque numeric internals of classical methods,
whose individual steps are not easily explained in
domain terms. An analyst can inspect the trace to check that the search respected
the clinical reasoning behind the targets and constraints, to detect
when the optimizer has latched onto an implausible region, and to
document and defend the modelling choices to collaborators or
reviewers. This traceability is particularly valuable in the
regulated health-technology-assessment context in which CEA models
are submitted, where modelling decisions must be transparent and
justifiable.

This paper introduces and evaluates an agentic calibration method
built on these ideas, and compares it on a 10-dimensional
anal cancer model against Nelder--Mead as a classical baseline
and BAxUS as a representative state-of-the-art BO method. Our
contribution is fourfold. First, we describe an LLM-driven
iteration loop with an explicit, auditable protocol suitable for
regulated CEA workflows, in which every iteration records both
the proposed parameter vector and a written rationale. Second, we
introduce a constrained variant that requires no additional
surrogate, only a textual description of the constraints. Third,
we report head-to-head benchmarks on a 10-dimensional anal cancer
model under both unconstrained and constrained calibration. 
Finally, we directly discuss the
trade-offs introduced by the LLM in the loop, including inference
cost, reproducibility, and the absence of formal convergence
guarantees.

% =============================================================
\section{Background}

This section reviews the main components relevant to our setting:
cost-effectiveness analysis as the application domain, Bayesian
Optimization and its high-dimensional and constrained variants as the
standard approach to calibration, and recent work on large language
models as potential in-context optimizers.

\subsection{Cost-effectiveness analysis}

CEA compares competing health
interventions by weighing incremental costs against
incremental health benefits, typically summarised as the incremental
cost-effectiveness ratio (ICER), the additional cost per unit of
health gained such as a quality-adjusted life year (QALY)
\citep{drummond2015,sanders2016}. When the relevant clinical events
unfold over long horizons or cannot be directly observed in trials,
these quantities are estimated with simulation models, most commonly
state-transition Markov cohort models or microsimulations, that
project disease progression, costs, and outcomes under each strategy
\citep{caro2012,siebert2012}.

The parameters governing these models, for instance transition
probabilities between health states, are often not directly measured but are taken
from the scientific literature or elicited from expert sources, each
of which carries its own intrinsic uncertainty. Because these initial estimates rarely
reproduce observed real-world outcomes, the model must be calibrated:
the free parameters are tuned until the model's outputs match a set of
observed targets, such as prevalence or mortality by age group
\citep{stout2009,vanni2011}. Because each candidate parameter vector
requires a full run of the simulation, and clinical knowledge
restricts plausible parameter combinations, calibration is
naturally cast as an expensive, constrained optimization problem.

\subsection{Bayesian Optimization}

Bayesian Optimization \citep{shahriari2016,garnett2023} is the
standard framework for optimizing expensive black-box objectives,
those for which each evaluation is costly and no gradient information
is available. Rather than searching the parameter space directly, it
maintains a probabilistic surrogate $p(f \mid \mathcal{D})$ of the
unknown objective, $f$, given the evidence gathered so
far, $\mathcal{D}$. The surrogate is often a Gaussian Process (GP), which
returns at each candidate point both a predicted mean and a
calibrated uncertainty. This uncertainty estimate lets
the loop decide where to sample next. At each step the method selects
the query
$x_{t+1} = \arg\max_x \alpha(x; \mathcal{D})$, where $\alpha$ is
an acquisition function such as Expected Improvement (EI) or Upper
Confidence Bound (UCB) \citep{shahriari2016}. The acquisition
encodes an explicit exploration--exploitation trade-off: high
mean improvement and high posterior uncertainty both increase the
score, so the loop balances refining the current best region against
probing regions it has not yet explored. The selected point is then
evaluated, the surrogate is updated, and the cycle repeats until the
evaluation budget is exhausted. This sample efficiency is what makes
BO attractive when each evaluation is computationally expensive.

\subsection{High-dimensional BO and BAxUS}

Standard BO degrades sharply as the dimension grows, because the
posterior uncertainty becomes nearly uniform across the search
space and the acquisition surface becomes dominated by boundary
effects \citep{wang2016rembo}. Several scalable variants address this issue.
In the CEA calibration context,
BO with additive kernels has been used for stepwise calibration
of simulation models, exploiting structure in how parameter
subsets affect the simulated outputs \citep{gomez2024}. SAASBO places a
sparse axis-aligned prior over the active dimensions
\citep{eriksson2021}. A separate
line of work uses local trust regions: TuRBO restricts the search
to trust regions whose size is modulated by the GP's local
lengthscales \citep{eriksson2019}, and BAxUS
\citep{papenmeier2022} extends this idea by adaptively embedding
the search into nested random subspaces, whose dimensionality is
increased only when there is evidence that the current
embedding has been exhausted.

\subsection{Constrained BO}

In the constrained setting, the standard approach is the
Probability of Feasibility (PoF) framework of
\cite{gardner2014}: each constraint $c_i(x) \le 0$ is modelled by
its own GP, and the ordinary acquisition $\alpha(x)$ is multiplied by
the surrogate's estimated probability that every constraint holds at $x$,
\begin{equation}
\alpha_{\text{c}}(x) = \alpha(x) \cdot \prod_{i} \underbrace{\Phi\!\left(\frac{-\mu_i(x)}{\sigma_i(x)}\right)}_{\Pr[\,c_i(x)\,\le\,0\,]}.
\end{equation}
Here $\mu_i(x)$ and $\sigma_i(x)$ are the posterior mean and standard
deviation of the $i$-th constraint GP at $x$, and $\Phi$ is the standard
normal CDF, so each factor is the probability that constraint $i$ is
satisfied. This product shrinks the acquisition of
likely-infeasible points toward zero.
This formulation is principled but introduces additional machinery:
one extra GP per constraint, an additional evaluation of each
constraint function on every queried point, and careful optimization of the
joint acquisition function. The approach scales poorly as the
number of constraints grows and the dimension increases,
since each additional GP suffers from the same lengthscale
estimation and conditioning issues that motivate high-dimensional
BO in the first place; specialised constrained variants such as
SCBO \citep{eriksson2021scbo} have been proposed to
address these limitations. Despite these developments, Bayesian
Optimization still requires increasingly complex modelling machinery in
such settings, motivating the exploration of alternative approaches.

\subsection{Large language models}

Large language models (LLMs) are the result of decades of
progress in statistical and computational approaches to language. Early
$n$-gram models gave way in the 2000s to language models
that learned distributed word representations, and the
introduction of the Transformer architecture in 2017
\citep{vaswani2017attention} removed the sequential bottleneck of
recurrent networks and enabled training at unprecedented
scale. Successive generations of pretrained models, such as BERT for
encoder-style understanding and the GPT family for generative
modelling, demonstrated that a single architecture, trained on
broad text corpora with a self-supervised objective, could be
adapted to a wide range of tasks. The release of
GPT-3 \citep{brown2020fewshot} marked a turning point by showing
that, beyond a certain scale, models acquire the ability to
perform new tasks from a handful of natural-language examples in
the prompt, with no parameter updates. This in-context learning
behaviour reframed LLMs as general-purpose programmable
components rather than as fixed text generators.

The impact on artificial intelligence has
been substantial. The combination of instruction tuning and
reinforcement learning from human feedback produced assistants
that reliably follow structured prompts and complex
instructions, and consumer-facing systems such as ChatGPT
brought conversational AI into mainstream use within months of
their release. More recent models devote part of their decoding
budget to explicit ``reasoning'' tokens before producing a final
answer, materially improving performance on tasks that require
multi-step arithmetic, planning or quantitative comparison.
Beyond natural-language tasks, LLMs and their multimodal
extensions are now routinely applied across science and
engineering: in software development, where they assist with
code generation, review and debugging; in the life sciences,
where they support literature synthesis, protein design and
molecular property prediction; in chemistry and materials, where
agentic systems plan and prioritise experiments
\citep{ramos2024review}; and in clinical and biomedical
informatics, where they are used for clinical decision support,
medical question answering and structured extraction from
electronic health records. The breadth of these applications
motivates the question taken up in this paper: whether the same
class of models can serve as a competitive optimizer inside a
quantitative scientific workflow such as CEA model calibration.

\subsection{LLMs as optimizers}

The ability of large language models to perform tasks specified
through a small number of natural-language examples
\citep{brown2020fewshot} has motivated their use as in-context
optimizers, where they propose candidate solutions from a textual
specification and feedback about previous trials
\citep{yang2024large,liu2024llmevopro}. This behaviour rests on
in-context learning, which has been characterised as a form of
implicit Bayesian inference: by conditioning on the examples in
its prompt, a pretrained Transformer effectively infers a latent
task and predicts accordingly \citep{xie2022icl_bayes}. When the inferred task
is optimization, the in-context history of trials plays the role
of conditioning data, while the next-token distribution acts as
the acquisition policy. From this perspective, an LLM
can be viewed as a soft acquisition function: a single in-context pass
implicitly weighs prior evidence and proposes a next point that
trades exploration against exploitation.

When an LLM is embedded in a loop of this kind, where it repeatedly
observes feedback, reasons about it, and acts to pursue a goal, it is
usually referred to as an agent, and the resulting system as agentic.
In contrast to a single question-and-answer exchange, an agent
operates over multiple steps, maintains a running context of what it
has tried and observed, and chooses each action in light of the
outcomes of its previous ones. In the calibration setting, the goal
is to minimise the calibration error, the observations are the
simulated outputs and residuals returned after each guess, and the
actions are the parameter vectors the model proposes.

% =============================================================
\section{Methods}
\label{sec:method}

\subsection{The agentic calibration method}
\label{subsec:agentic_method}

\subsubsection{Overview of the agentic calibration loop}

The central idea of the agentic method is to cast calibration as a
task that an LLM can perform by reasoning, much as a human modeller
would, rather than as a purely numerical search. A skilled modeller
tuning a model by hand does not treat it as a black box: they inspect
the current residuals, draw on their understanding of what each
parameter means and how it influences the outputs, and adjust their
next guess accordingly. The agentic method reproduces this workflow by
placing an LLM in the role of the optimizer and giving it the same
contextual information a modeller would rely on, while a harness, a small code layer that manages the
calibration loop, handles the simulation runs and keeps a record of the trajectory.

Concretely, the LLM is provided with the same information an
experienced human modeller would consult during a calibration
session: a description
of the model, the list of free parameters with their bounds and
their clinical meaning, the calibration targets, the history of
attempted guesses with their simulated outputs and the
corresponding errors, and, where applicable, the constraints.
Table~\ref{tab:llm_inputs} describes, for each of these
information items, the kind of information it represents and where
it comes from, using the anal cancer model of our experiments as a
running example; Appendix A provides further details
on these items, including the model description, the system prompt
through which they are conveyed to the LLM, and worked iteration
logs. At
every iteration, the LLM returns a new parameter vector together
with a short written rationale. The simulation is then run by the
calibration harness that mediates every model evaluation; the
resulting outputs and residuals are appended to the history that is
shown to the LLM on the next iteration. The loop terminates when
the iteration budget is exhausted, returning the best feasible
point observed during the run as the calibrated solution.

\begin{table}[ht!]
\centering
\caption{Information supplied to the LLM during the calibration
loop. For each item we describe the kind of information it
represents and where it comes from, illustrated with the
10-dimensional anal cancer model used in our experiments. The
history block is rebuilt every iteration from all evaluations
observed so far, and the constraints row applies only to the
constrained calibration setting.}
\label{tab:llm_inputs}
\small
\setlength{\tabcolsep}{4pt}
\begin{tabular}{p{0.20\textwidth}p{0.34\textwidth}p{0.34\textwidth}}
\toprule
\textbf{Information item} & \textbf{What it represents} & \textbf{Where it comes from (anal cancer model)} \\
\midrule
Model description &
A natural-language account of the model structure, health states
and dynamics, so the LLM can reason about how parameters affect the
outputs. &
A summary of the semestral Markov model of anal cancer natural
history (non-HSIL, HSIL and cancer stages), produced once from the
simulation source code by the code summarisation step. \\
\addlinespace
Free parameters and bounds &
The list of calibration variables and the admissible range of each,
defining the search space. &
The calibration specification: ten annual HSIL regression
probabilities (one per five-year age group, 30--34 to 75--79) with
their declared numeric bounds. \\
\addlinespace
Clinical meaning &
The semantic role of each parameter and the qualitative direction
in which it moves the outputs, acting as a prior over good values. &
Parameter metadata in the calibration specification: the
probability that a pre-cancerous HSIL lesion regresses, lower
values raising simulated cancer incidence. \\
\addlinespace
Calibration targets &
The observed quantities the simulated outputs must reproduce, and
against which residuals are computed. &
The age-specific anal cancer incidence curve drawn from the
published epidemiological literature used as the calibration target. \\
\addlinespace
History of guesses &
The trajectory of past evaluations: each proposed parameter vector
with its simulated outputs and the resulting errors, letting the
LLM learn from previous attempts. &
Recorded by the harness across iterations: each guess, the
simulated incidence per age group, and the residual feedback
(scalar error, RMSE, bias, per-group absolute and relative errors). \\
\addlinespace
Constraints &
Feasibility conditions every guess must satisfy, expressed in
domain terms. &
The clinical plausibility constraint from the specification: HSIL
regression probability non-increasing with age. \\
\bottomrule
\end{tabular}
\end{table}

Beyond the mechanics of the loop, the agentic method differs from
classical and BO-based optimizers in what it can use as input:
rather than scalar feedback alone, the LLM ingests natural-language
descriptions of the model, the parameters and the calibration
targets. The effective ``prior'' over good parameter vectors at
each iteration combines the parameter semantic metadata included
in the system prompt, a summary of the simulation model source
code produced once before calibration begins and described in
section~\ref{sec:code_summary}, and the model's own pretrained
knowledge of epidemiology and numerical optimization.

\begin{figure}[ht]
  \centering
  \includegraphics[width=\textwidth]{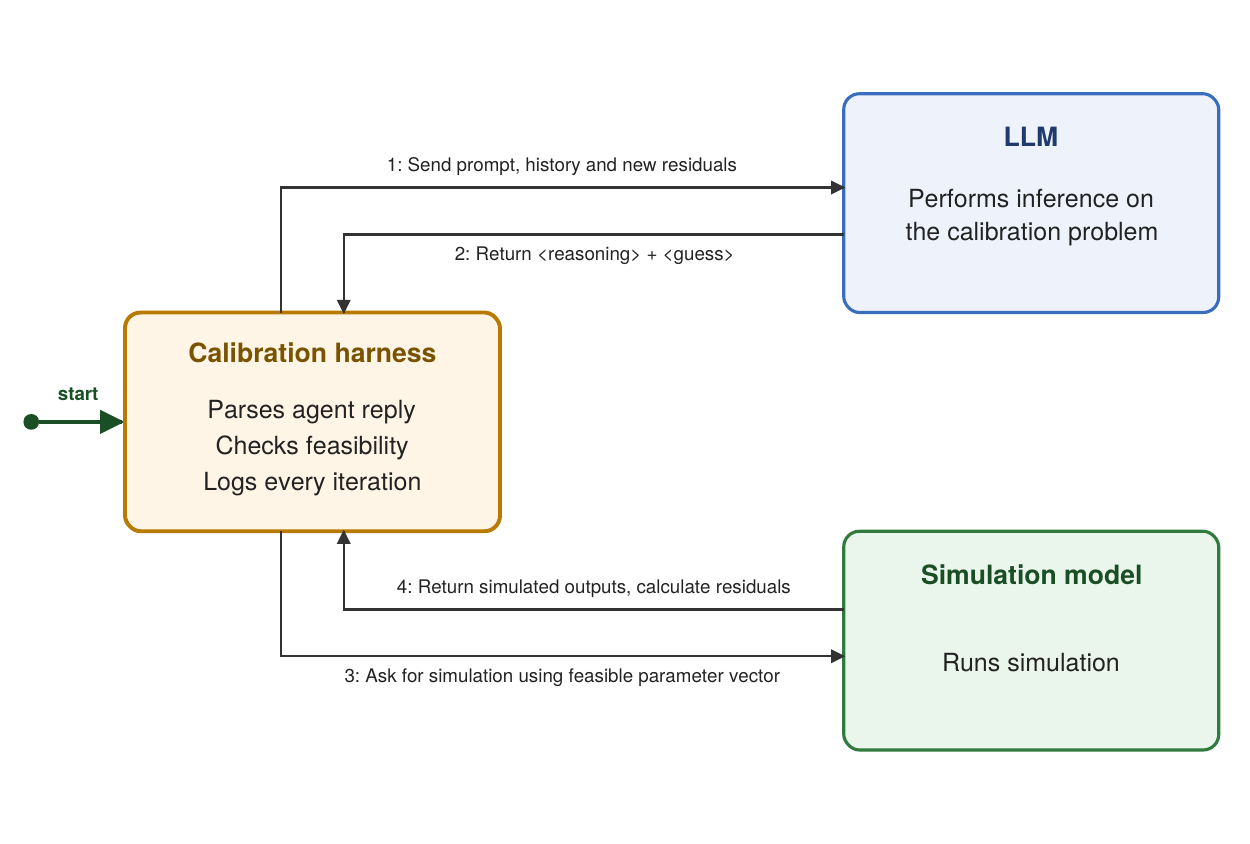}
  \caption{Agentic calibration loop. The harness drives the
  simulation model and the LLM through a fixed protocol. The LLM
  never invokes simulation calls directly; the harness mediates
  every evaluation, enforcing parameter bounds and recording the
  full trajectory.}
  \label{fig:architecture}
\end{figure}

The architecture is illustrated in Figure~\ref{fig:architecture}.

\subsubsection{The interaction protocol}

The LLM is required to produce outputs in a structured format with two
blocks: (i) a brief reasoning section, in which the model
describes what the previous residuals indicate and why the next
guess is expected to improve the fit; and (ii) a guess section, in
which the model returns the next parameter vector in a
machine-readable form. Two design choices are relevant to the
behaviour of the loop.

The first is that evaluations are mediated by the harness: the LLM
does not call the simulation model itself. The harness parses the
proposed guess, executes the simulation and supplies the resulting
outputs on the next iteration. This separation keeps every model
evaluation observable, replayable, and auditable by a non-technical
CEA reviewer, and it ensures that the cost of each iteration is
predictable.

The second is that the feedback is structured by group. At every
iteration, the LLM receives not only the scalar calibration error but
also a breakdown of residuals by clinically relevant stratum, such as
age group, including signed biases and both absolute and relative
deviations. This is the structured analogue of presenting a human
modeller with a calibration-target plot annotated by stratum. In
practice, the LLM uses these signals to localise which groups are
being missed and to make targeted parameter adjustments, rather than
treating the objective as an opaque scalar.

\subsubsection{Code summarisation}
\label{sec:code_summary}

Because the simulation model is a grey-box, its source code is made
available to the LLM in addition to the parameter and target
descriptions. As a pre-calibration step, the model source files
are concatenated, up to a configurable size limit and with
non-model directories excluded, and submitted in a single call to
an LLM, which is asked to produce a short summary describing how
each free calibration parameter is expected to affect the
simulated outputs. This summary is appended as a dedicated
section of the system prompt used by every subsequent iteration.
The summarisation step is performed once per calibration run and
adds a fixed, one-off inference cost. This step provides the LLM
with structural information about the model, derived from the source
code, that would otherwise be unavailable to black-box optimizers.
In deployments where the model source cannot be exposed to an LLM,
for example due to data-protection or intellectual-property
considerations, the summarisation step can simply be omitted,
and the calibration proceeds using only the parameter description
and the per-iteration log of evaluations.

\subsubsection{Constraint handling}

Adding constraints to the agentic method requires only extending the
calibration specification with a textual description of the
constraints, expressed in any form a domain expert would use:
``$p_{50} \le p_{40} \le \dots$'', ``no two adjacent transitions
may differ by more than a factor of two'', ``the cumulative
incidence in $[40,60]$ must not exceed $X$'', and so on. This
description is appended to the system prompt as a dedicated
Constraints section, with an explicit instruction that any guess
violating the constraints is to be considered invalid, even if it
would lower the error. Consequently, no additional surrogate is
fitted, no constraint-aware acquisition function is constructed,
and constraints can be modified at any time by editing the textual
description, with no changes to the optimizer required. This allows
constraints to be expressed at the level of domain reasoning rather
than as numerical functions required by standard optimizers.

Compared with the PoF framework, in which the modeller must
expose numeric constraint functions and the optimizer must fit an
additional GP per constraint, a combination that becomes
problematic as the number of constraints or the input dimension
grows \citep{eriksson2021scbo}, the agentic constraint
mechanism reduces the modeller's effort to a short textual
description.

\subsubsection{Robustness to hallucinations}
\label{sec:hallucinations}

A common concern when using LLMs in scientific workflows is the
possibility of spurious or ungrounded outputs (``hallucinations'')
that are plausible in form but not grounded in the underlying problem. Two
characteristics of the present design mitigate this risk in the
calibration setting.

The first is accountability through an explainable log. Each
iteration records the parameter guess together with its written
rationale, the resulting simulated outputs and per-stratum
residuals, and the corresponding update to the best-so-far value.
The trajectory is therefore fully auditable a posteriori, and any
proposal whose rationale is inconsistent with the residual pattern
or with the clinical constraints can be identified by the analyst
or by an external reviewer. When a calibration run fails to
converge to an acceptable error level, the same log enables a
post-mortem analysis: the iteration-by-iteration reasoning reveals
whether the agent was misinterpreting the residual pattern,
struggling to satisfy a particular constraint, or oscillating
between competing regions of the parameter space, providing
actionable guidance for prompt refinements.

The second is the limited agency granted to the LLM. It does not
execute the simulation, manipulate the constraint set, or modify the
calibration code. Every parameter vector is evaluated by the harness and checked
for feasibility before being incorporated into the trajectory.
Consequently, a hallucinated proposal at most yields a
sub-optimal evaluation in a single iteration; it cannot
compromise the integrity of the calibration loop nor produce a
result that violates the declared constraints.

\subsection{Experimental setup}

\subsubsection{Model and calibration settings}

We consider a 10-dimensional Markov model of anal
cancer natural history. The calibration parameters consist of ten
transition probabilities stratified by age group, corresponding
to the annual probability of regression of high-grade squamous
intraepithelial lesions (HSIL). Lower regression probabilities
imply a longer time spent in pre-cancerous states and therefore
higher simulated cancer incidence. As a result, the qualitative direction
in which each parameter should move given an observed residual is
often intuitive for a domain expert. Two settings are
evaluated. In the unconstrained calibration, the ten parameters
are allowed to vary within their natural ranges, with no clinical
plausibility constraints enforced. In the constrained calibration,
the same parameters are subject to a clinical monotonicity
constraint, namely that HSIL regression probability follow a
non-increasing trend across age groups. A full description of the
model, including its health states, transition structure, semestral
cycle, calibration target and role in the cost-effectiveness
analysis, is provided in Appendix A.

\subsubsection{Compared methods}

We compare three optimizers. The first is NM, the
classical derivative-free simplex method \citep{neldermead1965},
included as a widely used and easily deployed baseline; in the
constrained calibration, infeasible points are penalised by returning
$+\infty$. We restrict the classical baseline to NM because, in earlier
experiments comparing several derivative-free optimizers on this class
of calibration problems, it proved the most sample-efficient classical
method, clearly outperforming alternatives such as simulated annealing
and particle swarm optimization \citep{gomez2023ccia}.

The second is BAxUS, a high-dimensional BO method that adaptively
embeds the search in nested random subspaces. As a surrogate we use a GP
with an Automatic Relevance Determination (ARD) kernel and a Gamma
prior on the lengthscales, which discourages degenerate near-zero or
very large lengthscales, improving numerical stability when data
are scarce. For the constrained calibration, we enforce feasibility by
wrapping BAxUS in the PoF framework.

The agentic method follows the
protocol described in section \ref{subsec:agentic_method}, with the system prompt built
from the calibration specification. The optimizer is a locally
hosted reasoning LLM, namely Qwen3.6 27B \citep{qwen3technical},
used with 4-bit quantisation and
a context length of 128k tokens. It is configured with a per-iteration
reasoning budget of 2{,}048 tokens and a preset sampling
temperature of 0.05. The code summarisation step
(section~\ref{sec:code_summary}) is performed once at
initialisation using the same backend.

\subsubsection{Inference hardware}
\label{sec:hardware}

All agentic experiments were executed on a local workstation
equipped with two NVIDIA RTX~4000 GPUs, with 20\,GB of VRAM each.
Because inference is performed on local
hardware, no per-token charges are incurred. As a reference for
deployments that would rely on a frontier-model API, a rough
estimate based on the prompt and reasoning token counts produced
by this setup places the equivalent cost at approximately
$0.10$\,€ per iteration using Claude's Opus 4.7, or about $1.60$\,€ per calibration run.
A typical CEA project requires a moderate number of calibration runs,
so both options remain compatible with standard budgets.

\subsubsection{Evaluation budgets}

To reflect the very different per-iteration
costs of the methods, the following evaluation budgets are used. NM uses the
standard convergence criterion of the simplex method, which in
practice consumes from a few hundred to several thousand model
evaluations. BAxUS is given 10 initial random samples and 100 BO
iterations, for a total of 110 model evaluations. The agentic method
uses one initial-guess evaluation and 15 LLM iterations, for a
total of 16 model evaluations.

Each method is run ten times under both calibration settings,
varying the random seed across runs. We report
the median, mean, and standard deviation of the best feasible
error attained.

% =============================================================
\section{Results}
\label{sec:results}

Across both calibration settings, the agentic method reduces the
calibration error by two to three orders of magnitude compared to
Nelder--Mead. Compared to BAxUS, it achieves comparable or better
solution quality while using roughly one-seventh as many model
evaluations. Table~\ref{tab:summary} reports the best feasible error
reached by each method, together with the evaluation count and
wall-clock time, aggregated over the ten runs under both calibration
settings on the 10-dimensional anal cancer model. We report calibration accuracy first, then a qualitative
inspection of the calibrated solutions, the computational performance
in wall-clock time, and finally the monetary cost.

\begin{table}[ht!]
\centering
\small
\setlength{\tabcolsep}{4pt}
\begin{tabular}{llrr r@{\,$\times$\,}l r@{\,$\times$\,}l r@{\,$\times$\,}l}
\toprule
& & & & \multicolumn{6}{c}{\textbf{Best error}} \\
\cmidrule(lr){5-10}
\textbf{Setting} & \textbf{Method} & \textbf{Evals} & \textbf{Time (s)} & \multicolumn{2}{c}{\textbf{Median}} & \multicolumn{2}{c}{\textbf{Mean}} & \multicolumn{2}{c}{\textbf{SD}} \\
\midrule
Unconstrained & NM       & $486$    & $703$     & $2.87$ & $10^{-8}$  & $3.01$ & $10^{-8}$  & $1.01$ & $10^{-8}$  \\
              & BAxUS    & $110$    & $573$     & $1.34$ & $10^{-9}$  & $1.34$ & $10^{-9}$  & $1.97$ & $10^{-11}$ \\
              & Agentic  & $16$     & $3{,}981$ & $5.03$ & $10^{-11}$ & $8.27$ & $10^{-11}$ & $1.03$ & $10^{-10}$ \\
\midrule
Constrained   & NM       & $21{,}917$ & $3{,}129$ & $3.45$ & $10^{-7}$  & $3.54$ & $10^{-7}$  & $2.70$ & $10^{-8}$  \\
              & BAxUS    & $110$    & $1{,}127$ & $1.33$ & $10^{-9}$  & $1.30$ & $10^{-9}$  & $6.77$ & $10^{-11}$ \\
              & Agentic  & $16$     & $5{,}133$ & $1.56$ & $10^{-9}$  & $1.65$ & $10^{-9}$  & $2.77$ & $10^{-10}$ \\
\bottomrule
\end{tabular}
\caption{Best feasible error, evaluation count and wall-clock time
on the 10-dimensional anal cancer model. The median, mean and
standard deviation of the error, together with wall-clock times,
are reported across ten runs. For BAxUS and the agentic method the evaluation count is
fixed by the budget, while for NM, whose simplex terminates by its
own convergence criterion, the mean number of evaluations is reported.}
\label{tab:summary}
\end{table}

\begin{figure}[ht!]
  \centering
  % Generated by docs/agentic_article/figures.R
  \includegraphics[width=\textwidth]{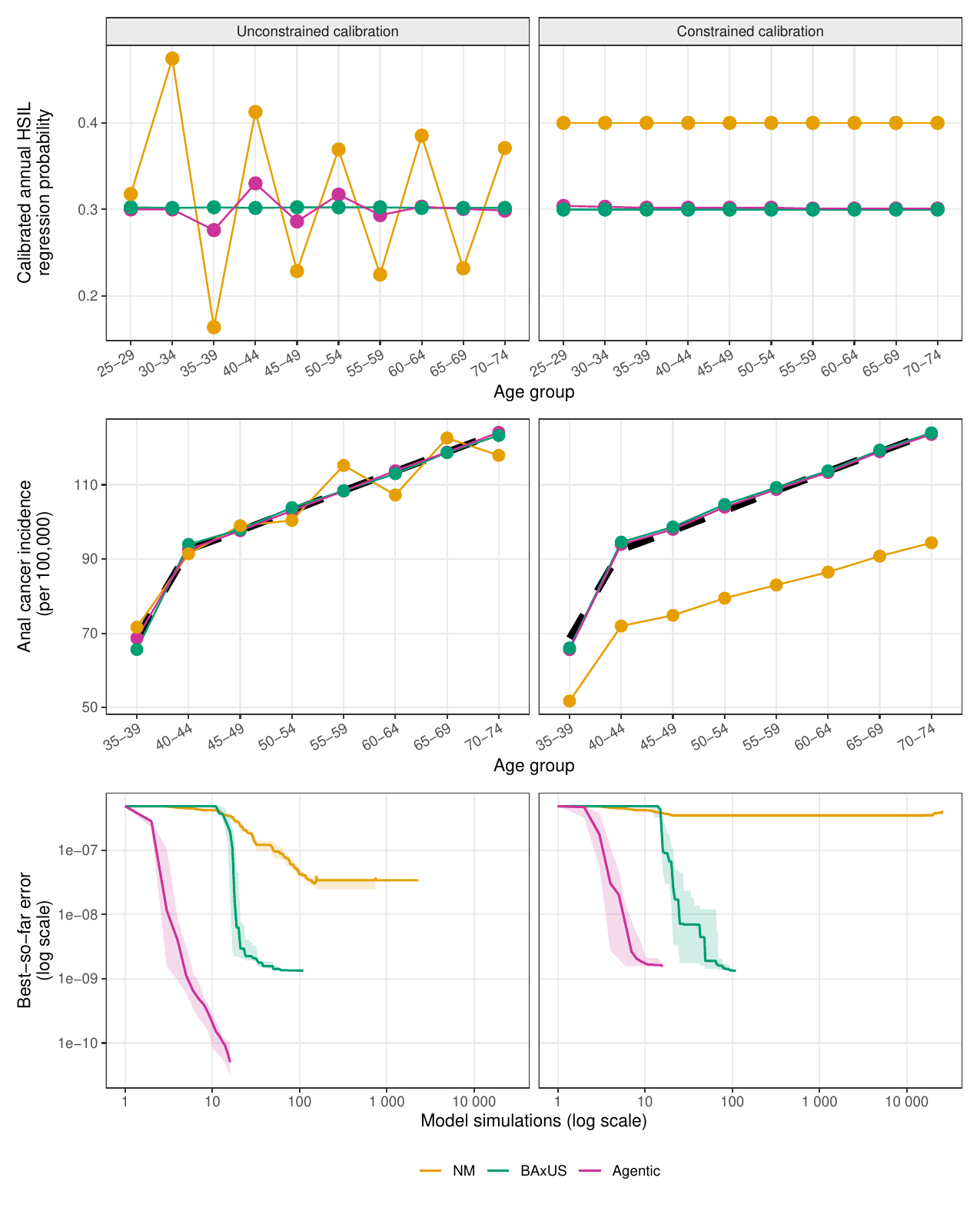}
  \caption{Per-method comparison on the 10D anal cancer model, for
  the NM, BAxUS and agentic methods. Columns: unconstrained (left)
  and constrained (right) calibration. The solid black line in the
  middle row is the observed calibration target. In the bottom row,
  the shaded area corresponds to the inter-quartile range (IQR) of ten calibrations.}
  \label{fig:results}
\end{figure}

Figure~\ref{fig:results} summarises the comparison across the three
methods in three rows. The top row presents each method's
best-found parameter vector, namely the calibrated annual HSIL
regression probabilities across the age strata. The middle row shows the corresponding
model output obtained with those probabilities: the simulated anal
cancer incidence (coloured lines) against the
observed calibration target (dashed black line). The bottom row shows
the overall error of the curves above, as the best-so-far calibration
error per iteration across ten runs. Note how, under unconstrained calibration, the calibrated
probabilities are not coherent and oscillate across adjacent age strata
(top-left panel), even though the resulting model output still fits the
target closely; the constrained scheme removes this oscillation by
enforcing the clinical plausibility constraint.

\subsection{Calibration accuracy}

Under unconstrained calibration, the agentic method reaches a median best error approximately
$27\times$ lower than BAxUS and nearly three orders of magnitude
lower than NM, despite using about one-seventh of the evaluations
of BAxUS and a small fraction of those of NM. The per-iteration convergence
behaviour is shown in the left column of the bottom row of
Figure~\ref{fig:results}.

Under constrained calibration, the agentic method is
within the same order of magnitude as BAxUS,
and both clearly outperform NM, which does not reach the
$10^{-7}$ level despite consuming on the order of $10^4$
evaluations (Figure~\ref{fig:results}, bottom row, right column).
Notably, the agentic method achieves this performance using only a
textual description of the constraints, without any additional
constraint-handling machinery. This difference in modeller-facing
complexity is not reflected in wall-clock time, but becomes evident
when specifying constraints, which requires substantially more
engineering effort in BO-based methods, despite the substantially
smaller evaluation budget of the agentic method.
The agentic standard deviation is somewhat larger than in the
unconstrained case in absolute terms, but all ten runs converge
within a factor of two of the BAxUS median, indicating that the
constrained variant remains robust. This increased variance likely
reflects the tighter feasible region induced by the constraints, which
reduces the margin for corrective adjustments.

\subsection{Computational performance}

Most of the wall-clock time of the agentic method is spent on LLM
inference rather than on simulation calls. On our infrastructure, a full
agentic run takes roughly 4--7$\times$ as long as a BAxUS run on the
same problem.

Under constrained calibration the picture changes for the
classical baseline method as well. Because NM has no native
constraint-handling mechanism and must reject every infeasible
candidate it proposes, its evaluation count rises from $486$ in the
unconstrained setting to $21{,}917$ when constraints are enforced, a
roughly $45\times$ increase. Almost all of these additional
evaluations are spent on infeasible candidates that are discarded,
which is why NM ends up slower than BAxUS despite the latter's
surrogate-model overhead. The agentic method is also slower than BAxUS in this
setting, for the same inference-bound reason as in the
unconstrained case. This highlights that the overall performance
depends on the relative cost of simulation and inference.

% =============================================================
\section{Discussion}

The agentic method is particularly well suited to several
recurring situations in CEA calibration. A first situation is that
of very small evaluation budgets: a single simulation can take
minutes, in which case even 100 evaluations may be impractical,
and the structured per-group residual feedback used by the agentic
method enabled convergence in approximately fifteen iterations on
the 10-dimensional anal cancer model considered.

A second situation is that of grey-box problems with interpretable
parameters. Where the parameters and targets carry a clinical
interpretation, an experienced modeller can reason qualitatively
about which parameter adjustment a given residual pattern requires; for
instance, increasing the annual HSIL regression probability
reduces simulated cancer incidence in the affected age group. The
agentic method can leverage the same intuition by ingesting the
natural-language description of the model and parameters, whereas
derivative-free and BO-based optimizers must rediscover these
relationships from evaluations alone.

A third situation involves frequently changing constraints: calibration is often iterative for the analyst, with new plausibility constraints added as the
team learns more about the model's behaviour, and updating a
textual description is far less costly than re-engineering an
optimizer. A further practically important comparison, not
captured by standard performance metrics, concerns the
modeller-facing effort required to add a new clinical constraint.
Table~\ref{tab:constraint_effort} summarises the changes each
method requires; an extended version that details how the specific
constraint used in our experiment is encoded under each method is
given in Appendix B.

A fourth situation arises when an auditable rationale
is required. As discussed in section~\ref{sec:hallucinations},
each iteration produces a written justification of the proposed
parameter update, which can be retained as an artefact for
regulatory dossiers and peer review and used for post-mortem
analysis when a run fails to converge. Classical and BO-based
optimizers do not produce a directly comparable explanation of
their proposals.

\begin{table}[ht!]
\centering
\begin{tabular}{lp{10cm}}
\toprule
\textbf{Method} & \textbf{Required change} \\
\midrule
NM              & Modify the calibration objective to return
                  $+\infty$ on infeasible points. No surrogate or
                  acquisition function to update. \\
BAxUS (PoF)     & Implement an explicit numerical constraint
                  function whose sign encodes feasibility; one
                  additional GP per constraint is fitted and a
                  constrained acquisition is jointly optimized.
                  Care is required in choosing the scale of the
                  constraint function. \\
Agentic         & Extend the textual description of the
                  constraints in the calibration specification.
                  No changes to the optimizer are required. \\
\bottomrule
\end{tabular}
\caption{Modeller-facing changes required to add a new clinical
constraint, by method.}
\label{tab:constraint_effort}
\end{table}

The agentic method also involves a few practical considerations worth
keeping in mind. The first is inference cost, which amounts to a
trade-off between money and time.
LLM inference dominates the time per iteration, and on our hardware
the loop is about 4--7$\times$ slower than BAxUS. The size of this gap
depends on the computing resources available: more capable GPUs, or a
hosted API such as Claude or GPT, make inference faster and can bring
the loop level with or ahead of BAxUS, at a higher monetary cost. Once
inference is fast enough, the per-iteration time is instead dominated
by the simulation model itself, which is the common regime in practice,
especially when each simulation takes minutes to run.
Either way the method is scalable, and, as discussed in
section \ref{sec:hardware}, the total monetary cost stays modest for
both local and API deployments at the scale of a typical CEA project.

A second consideration is more theoretical than practical: the absence
of formal convergence guarantees. NM and BO both offer at least partial
theoretical guarantees of progress under standard assumptions, whereas
the agentic method's behaviour depends on the model, the prompt, and the
decoding parameters. In practice, however, the methodology proposed here
is precisely what guides the LLM towards convergence: the structured
calibration history, the explicit objective, and the iterative feedback
loop steer the model's choices in a consistent direction. Accordingly, in
our experiments performance was consistent, with only somewhat higher
variance under constraints, and this variance can plausibly be reduced
further through modest prompt refinement or a larger per-iteration
reasoning budget.

A third point to keep in mind when implementing the method concerns
reproducibility. An NM or BO run with a fixed random seed is exactly
reproducible on the same machine. An LLM-driven loop is reproducible
only insofar as the inference backend provides deterministic sampling
and keeps the model weights fixed. The
local-backend results reported here reproduce across runs under the
same configuration; an API deployment would additionally require the
provider to pin the model version to offer comparable guarantees.

A fourth consideration is that the effectiveness of the method depends on
the LLM possessing the domain knowledge required to reason about the model
and its parameters, since part of the available knowledge at each iteration is
the model's own pretrained understanding of the application domain
(section~\ref{subsec:agentic_method}). The anal cancer model used here
involves parameters and targets whose qualitative behaviour is well
represented in the general epidemiological and clinical literature on which
the LLM was pretrained. More complex or specialised domains, where the
relevant relationships are rare, proprietary, or absent from public
training corpora, may exceed the model's pretrained knowledge. In such
cases, fine-tuning the LLM on domain-specific material, so that it acquires
the knowledge needed for the agentic calibration to be effective, may be
necessary.

Finally, from a conceptual perspective, the agentic method can be
related to Bayesian Optimization and naturally suggests a range of
possible extensions. The agentic method can be viewed as a BO
loop in which the surrogate is the LLM's representation of the calibration
history and the acquisition function is the LLM's implicit rule for
choosing the next point. The analogy is informal, as there is no
closed-form posterior, no explicit balance parameter, and no
convergence guarantee, but it is useful. Exploitation is driven by the
running best value, the per-group residual signs, and prompt
instructions that favour refinement once a promising region is found.
Exploration arises both implicitly, from the model's sampling
temperature, and explicitly, from the instruction to take larger steps
in early iterations. Constraints are handled in the prompt, with the
LLM asked to justify that a guess is feasible before proposing it.

The closest framing in the BO literature is the recent line of
work on LLM-guided acquisition selection
\citep{liu2024llmevopro,song2024position}, in which the LLM
chooses which acquisition function to use at each step within an
otherwise standard, black-box BO loop. Our setting differs in two
ways: the LLM plays the role of the acquisition function rather
than choosing among predefined ones, and the problem is treated
as grey-box rather than black-box, allowing the LLM to
incorporate the textual description of the simulation model and the
clinical meaning of the parameters into its proposals.

These observations suggest a continuum of designs in which the
agentic method is the limit case where the surrogate is entirely
implicit. Intermediate designs constitute natural extensions of
this work, for example: (i) a small GP surrogate fitted in
parallel and used to validate the LLM's proposals before
evaluation; (ii) the LLM selecting among a fixed set of
acquisition functions at each step, in the spirit of
\cite{song2024position}; or (iii) the LLM generating candidates
inside a trust region defined by a TuRBO-style state
\citep{eriksson2019} to bound exploration explicitly. We consider
these promising directions for follow-up work.

% =============================================================
\section{Conclusions}

We presented an agentic calibration method for simulation models
in cost-effectiveness analysis. The method drives an LLM through
a structured iteration loop in which the model proposes parameter
vectors and a calibration harness mediates every simulation call.
On a 10-dimensional anal cancer model, the agentic method achieved
comparable or better solution quality than BAxUS under
both unconstrained and constrained calibration, while requiring
fewer model evaluations, indicating improved sample efficiency.
Constraint handling is achieved through a simple textual
description, avoiding the additional modelling machinery required by
BO-based approaches. By treating
calibration as a grey-box problem, the method leverages the
natural-language description of the model and parameters, an
information source that classical and BO-based optimizers cannot
directly exploit. The main trade-off introduced by this approach lies
in runtime: LLM inference increases the per-iteration cost, and the
method is slower than BAxUS under the experimental setup considered.
However, the reduced number of model evaluations makes the approach
particularly suitable when simulation time is the dominant
computational cost, a common situation in cost-effectiveness
modelling. A complementary limitation is the absence of formal
convergence guarantees, in contrast to classical and BO-based methods.
Beyond performance, the method produces a human-readable and auditable
optimization trace, which can facilitate model validation,
communication, and regulatory review in health technology assessment
settings.
Taken together, the agentic method provides a practical and
flexible alternative for model calibration, particularly in settings
with limited evaluation budgets and evolving clinical constraints.

% =============================================================

\section*{Competing interests}
The authors declare that they have no competing interests.

% =============================================================
\bibliographystyle{unsrtnat}

\end{document}

\appendix
\setcounter{section}{0}
\renewcommand{\thesection}{\Alph{section}}
% Number appendix figures as S1, S2, ...
\setcounter{figure}{0}
\renewcommand{\thefigure}{S\arabic{figure}}

\section*{Appendix}

This appendix provides the supporting material for the calibration
case study. Appendix~\ref{app:model} describes the anal cancer
simulation model used throughout the experiments, including its
structure, the calibrated parameter and the calibration target.
Appendix~\ref{app:constraint_encoding} then contrasts how the
clinical plausibility constraint of the constrained experiment is
encoded under the agentic method and under the Probability of
Feasibility (PoF) baseline.
Appendices~\ref{app:system_prompt}--\ref{app:summary_prompt}
reproduce verbatim the three prompts used by the agentic method: the
system prompt that drives the iteration loop
(Appendix~\ref{app:system_prompt}), the constraint section appended
to it in the constrained setting
(Appendix~\ref{app:constraint_prompt}), and the code summarisation
prompt run once before calibration begins
(Appendix~\ref{app:summary_prompt}). Finally,
Appendices~\ref{app:log_excerpt_unconstrained}
and~\ref{app:log_excerpt_constrained} give excerpts of representative
per-iteration logs for the unconstrained and constrained runs,
respectively. Curly-brace placeholders in the prompts (for example
\texttt{\{scheme description\}}) are filled at run time from the
calibration specification.

% =============================================================
\section{The anal cancer simulation model}
\label{app:model}

\subsection*{Goal and context}

The case study calibrates a natural-history model of anal cancer in
HIV-positive men who have sex with men (MSM), a population at
markedly elevated risk of the disease. The model underpins a
cost-effectiveness analysis of anal cancer screening strategies, and
is calibrated so that its simulated incidence reproduces observed
epidemiology before any strategy is evaluated.

\subsection*{Model structure}

The model is an age-stratified Markov state-transition model that
follows a cohort of HIV-positive MSM through three clinically
meaningful stages: a disease-free state, a pre-cancerous state
corresponding to high-grade squamous intraepithelial lesions
(HSIL), and anal cancer. Disease-free individuals may acquire HSIL;
HSIL may either regress back to the disease-free state or progress
to cancer; and cancer leads to survival or death. The balance
between HSIL regression and progression therefore governs how much
of the cohort eventually reaches cancer.

\begin{figure}[ht!]
  \centering
  \includegraphics[width=\textwidth]{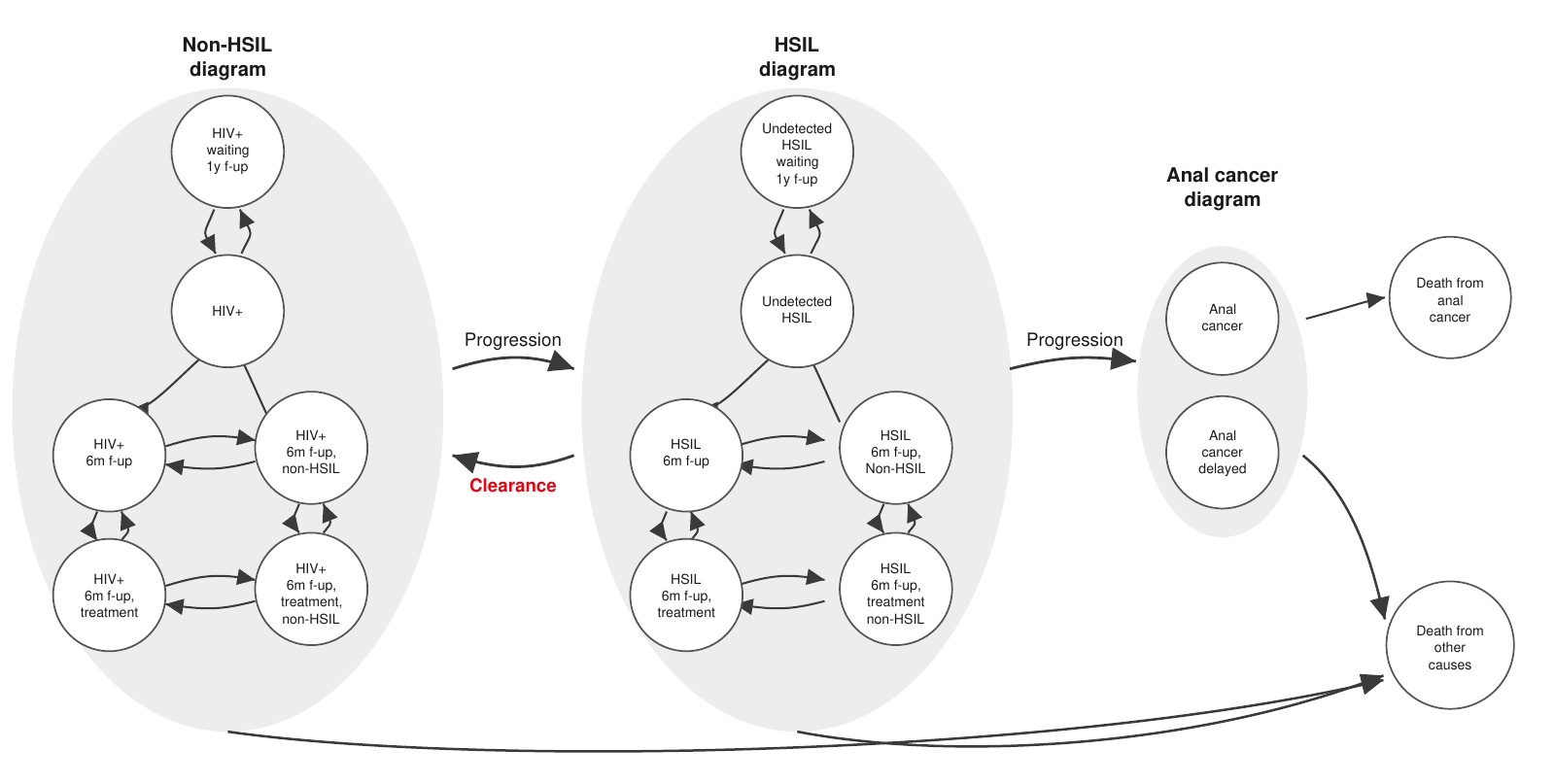}
  \caption{State-transition diagram of the anal cancer
  natural-history model. States are grouped into the non-HSIL
  (disease-free), HSIL (pre-cancerous) and anal cancer stages;
  arrows between groups denote progression and clearance (the
  latter, highlighted in red, is the HSIL regression that the
  calibration tunes), and every state may transition to death from
  other causes.}
  \label{fig:anus-states}
\end{figure}

Figure~\ref{fig:anus-states} shows the state-transition diagram.
The three stages are grouped into the non-HSIL (disease-free), HSIL
(pre-cancerous) and anal cancer blocks. The fine-grained states
within each block correspond to screening and follow-up status,
while the inter-block arrows are the natural-history transitions of
interest: progression towards cancer and clearance (HSIL
regression) back to the disease-free stage. All states may exit to
death from other causes.

\subsection*{Calibrated parameter}

Calibration tunes the annual probability of HSIL regression, which
has a clear and monotone effect on the output: a lower regression
probability keeps individuals longer in the pre-cancerous state and
thus raises simulated cancer incidence, while a higher value lowers
it. This probability is stratified by age, yielding the
ten-dimensional parameter vector used in the experiments, and its
clinical interpretation is also what makes the plausibility
constraint (regression non-increasing with age) natural to state.

\subsection*{Calibration target and objective}

The target is the age-specific anal cancer incidence in
HIV-positive MSM reported by the meta-analysis of Clifford et
al.\ (2021).\footnote{G.~M. Clifford et al., ``A meta-analysis of
anal cancer incidence by risk group: Toward a unified anal cancer
risk scale,'' \textit{International Journal of Cancer}, 2021.} At
each evaluation the simulator returns the simulated anal cancer
incidence per age group, and the objective is its squared deviation
from the target curve. The per-group residuals fed back to the
agent each iteration are the components of this objective broken
down by age group, as illustrated in the iteration logs of
Appendices~\ref{app:log_excerpt_unconstrained}
and~\ref{app:log_excerpt_constrained}.

% =============================================================
\section{Encoding the experiment constraint: agentic vs.\ PoF}
\label{app:constraint_encoding}

This section contrasts, side by side, how the single clinical
constraint used in the constrained experiment is introduced under the
agentic method and under the regular Probability of Feasibility (PoF)
framework used to wrap the BAxUS baseline. The constraint is the
monotonicity (clinical plausibility) requirement that the annual HSIL
regression probability be non-increasing with age. For the
ten-dimensional parameter vector
$(p_{30\text{-}34}, p_{35\text{-}39}, \dots, p_{75\text{-}79})$ this is
the conjunction of nine adjacent-pair inequalities
\[
  p_{30\text{-}34} \ge p_{35\text{-}39} \ge \dots \ge p_{75\text{-}79},
  \qquad\text{i.e.}\qquad
  c_i(p_1, \dots, p_{10}) = p_{i+1} - p_i \le 0,\quad i = 1,\dots,9 .
\]

Under the agentic method the same requirement is conveyed as a single
domain-level sentence in the Constraints section of the system prompt
(Appendix~\ref{app:constraint_prompt}); under PoF it must be expressed
as the nine explicit numerical constraint functions $c_i$ above, each
fitted with its own Gaussian process (GP) surrogate and combined into
the constrained acquisition of
Equation~\eqref{eq:pof} (reproduced here for convenience),
\begin{equation}
\alpha_{\text{c}}(x) = \alpha(x) \cdot \prod_{i=1}^{9}
  \Phi\!\left(\frac{-\mu_i(x)}{\sigma_i(x)}\right),
\label{eq:pof}
\end{equation}
where $\mu_i$ and $\sigma_i$ are the posterior mean and standard
deviation of the $i$-th constraint GP. Table~\ref{tab:constraint_encoding}
summarises the resulting differences for this experiment.

\begin{table}[ht!]
\centering
\caption{How the experiment's monotonicity constraint (HSIL regression
probability non-increasing with age) is added under each method.}
\label{tab:constraint_encoding}
\footnotesize
\begin{tabular}{p{3.0cm} p{5.4cm} p{5.4cm}}
\toprule
\textbf{Aspect} & \textbf{Agentic method} & \textbf{Regular method (PoF)} \\
\midrule
What the modeller writes &
One natural-language sentence appended to the Constraints section of
the prompt: ``HSIL regression probabilities should decrease or remain
constant with age.'' &
Nine numerical constraint functions $c_i(p_1, \dots, p_{10}) = p_{i+1} - p_i \le 0$, one
per adjacent age-stratum pair, returning a scalar whose sign encodes
feasibility. \\
\addlinespace
Representation &
Free-text, domain-level; the strata are referred to collectively, not
enumerated. &
Explicit algebraic functions over the parameter vector; every pair must
be written out. \\
\addlinespace
Number of constraint objects &
$0$ explicit functions (one sentence covers all strata). &
$9$ constraint functions. \\
\addlinespace
Auxiliary models &
None. &
One additional GP fitted per constraint ($9$ extra GPs), each requiring
hyperparameter estimation alongside the objective surrogate. \\
\addlinespace
Enforcement mechanism &
The harness evaluates the nine inequalities on each proposed vector and rejects infeasible proposals before
the simulator is invoked; the agent is additionally asked to justify
feasibility in its reasoning block. &
The acquisition is multiplied by the product of per-constraint
feasibility probabilities (Equation~\eqref{eq:pof}) and the resulting
constrained acquisition is jointly optimised, steering proposals toward
the feasible region. \\
\addlinespace
Tuning required &
None. &
The scale of each constraint function and the GP hyperparameters must
be chosen; conditioning degrades as the number of constraints and the
input dimension grow. \\
\addlinespace
Changing the constraint &
Edit the sentence in the calibration specification; no code or
optimiser changes. &
Re-implement the constraint functions and refit the corresponding GPs. \\
\bottomrule
\end{tabular}
\end{table}

% =============================================================
\section{System prompt for the iteration loop}
\label{app:system_prompt}

The following prompt is sent as the first message of the iteration
loop, instructing the agent on the problem, the reply format and
the meaning of each field in the per-iteration feedback. The code summarisation is
prepended to this prompt as an additional ``Model summary'' section.

\begin{lstlisting}[style=prompt]
You are a calibration agent for a Markov-based health economic
model.
Goal: pick parameter values that make the model match the target.

## Scheme
{scheme description}

## Free parameters (respect bounds)
{
  "<param_1>": { "type": "number", "minimum": ..., "maximum": ... },
  ...
}

## Calibration target
{ "<group_1>": <number>, "<group_2>": <number>, ... }

## How to respond
On every turn, return BOTH of these blocks, in this order:
  1. <reasoning>...</reasoning>  -- 1-4 sentences explaining what
     you observed in the previous results and why this new guess
     should improve the fit.
  2. <guess>{...}</guess>        -- a single JSON object with every
     parameter name above as a key and a number within its bounds
     as the value. The <guess> block must contain ONLY the JSON
     object (no comments, no prose).
Do NOT call any tool. The application runs the simulation for you
and sends back the simulated output and error on the next turn.

## Result fields you will receive each iteration
- error                : the scalar objective the simulator reports
                         (lower is better).
- rmse                 : sqrt of mean squared (sim - target) over
                         groups.
- bias                 : mean(sim - target) over all groups
                         (signed).
- max_abs_error        : largest |sim - target| across groups.
- abs_error_by_group   : |sim - target| per group (magnitude of
                         miss).
- bias_by_group        : sim - target per group (signed; positive =
                         overestimate).
- rel_bias_by_group    : (sim - target) / target per group
                         (signed, relative).
- simulated, target    : raw simulated and target values per group.

Strategy:
- Read abs_error_by_group to spot which groups are missed the most.
- Read bias_by_group / rel_bias_by_group to spot direction:
  positive => overestimating that group (consider lowering
  parameters that drive it up); negative => underestimating.
- Try to reduce RMSE each iteration; large moves early, refine
  later.
\end{lstlisting}

% =============================================================
\section{Constraint prompt extension}
\label{app:constraint_prompt}

In the constrained calibration setting the system prompt is
extended with the following section. The constraint description is
supplied by the calibration specification and is reproduced
verbatim in the prompt.

\begin{lstlisting}[style=prompt]
## Constraints (every guess MUST satisfy these)
{constraint description:
  The annual HSIL regression probability must decrease
  monotonically with age:
      p_30_34 >= p_35_39 >= p_40_44 >= ... >= p_75_79.
  All values must remain within their declared bounds.
}
Any guess that violates the constraints above is invalid even if it
would lower the error. Use the <reasoning> block to justify how your
guess stays within the constraints, and only then emit the <guess>
block.
\end{lstlisting}

Compliance is enforced by the harness independently of the agent's
self-assessment. Proposals that fail the feasibility check are
rejected before the simulation model is invoked.

% =============================================================
\section{Code summarisation prompt}
\label{app:summary_prompt}

The pre-calibration code summarisation step uses a separate
system prompt, in which the agent is shown the simulation source
code and is asked to produce a short factual summary structured as
one bullet per base parameter. This step is run once before the
iteration loop begins, and its output is prepended to the iteration
loop prompt as a ``Model summary'' section. The prompt is reproduced
below.

\begin{lstlisting}[style=prompt]
You are analyzing the source code of a Markov-based health economic
simulation model that will be calibrated by a downstream agent.
You will receive (a) the free calibration parameters with their
bounds, (b) how those parameters are grouped by base parameter
(stratified parameters share one group across strata), (c) the
calibration target, and (d) the model source code.

Produce a factual summary as a Markdown bullet list with EXACTLY
ONE top-level bullet per base parameter group ({n_groups} groups
total). Use the base parameter name as the bullet header, then
describe in 2-4 sentences:
  - what the parameter represents in the simulation;
  - the qualitative direction in which raising or lowering it
    moves the simulated output, and which target component(s) /
    stratum / age groups it most affects;
  - for stratified groups, note whether all strata behave the same
    way or whether some strata are more sensitive than others;
  - any couplings, trade-offs, or shared effects with other
    parameter groups, and any obvious sensitivities, saturation
    effects, or boundary behaviors.

Format strictly as:
- **<base_param_name>**: <description>
- **<base_param_name>**: <description>
...

Do NOT add introductory text, headers, code blocks, greetings,
caveats, or meta commentary. Do NOT emit more than one bullet per
base parameter group. Do NOT split stratified parameters into
separate top-level bullets -- keep them grouped under their base
name.
\end{lstlisting}

% =============================================================
\section{Representative iteration log: unconstrained calibration}
\label{app:log_excerpt_unconstrained}

The following excerpt is taken verbatim from a log produced during
the unconstrained calibration of the anal cancer model. The summary
of the simulation model was emitted by the code summarisation step and
prepended to the iteration loop prompt. The reasoning blocks
reproduced below were emitted by the agent at the first two
iterations and at the final iteration of the run; the residual
tables are the structured feedback that the harness returned after
each proposal. The scalar error decreases from
$2.82\times10^{-7}$ at iteration~1 to $1.71\times10^{-8}$ at
iteration~2, and reaches $4.91\times10^{-11}$ at the final
iteration, illustrating how the agent uses signed bias to redirect
the parameter trajectory between successive iterations and how the
per-stratum corrections become progressively finer-grained as the
run converges.

\subsection*{Model summary (output of the code summarisation step)}

\begin{lstlisting}[style=prompt]
- **p_hsil_regression_annual**: This parameter represents the
  annual probability that an individual with high-grade squamous
  intraepithelial lesion (HSIL) regresses to a non-HSIL state,
  directly governing the average duration spent in the
  precancerous compartment. Raising this probability shrinks the
  at-risk HSIL pool and decreases simulated anal cancer incidence
  across all calibrated age groups (y40_44 through y75_79), while
  lowering it increases incidence. Older strata tend to be more
  sensitive to changes because they have accumulated higher HSIL
  prevalence over time, making the regression rate a stronger
  determinant of their long-term cancer risk. As the sole
  calibration lever, it must simultaneously fit an upward-sloping
  age-specific incidence curve, creating a structural trade-off
  where regression rates generally need to decline with age to
  match rising targets; values near the [0, 1] bounds can trigger
  saturation effects where cancer incidence flattens at a floor or
  ceiling, reducing gradient information for the optimizer.
\end{lstlisting}

\subsection*{Iteration 1}

Reasoning block emitted by the agent:

\begin{lstlisting}[style=prompt]
The model consistently underestimates cancer incidence across all
calibrated age groups by approximately 23-24%, indicating that the
annual HSIL regression probability is currently too high. Lowering
these regression rates will increase the average time individuals
spend in the precancerous HSIL state, thereby increasing the
probability of progression to cancer and raising simulated
incidence. Since the relative bias is nearly uniform across ages,
a global reduction in all regression parameters from 0.4 to 0.25
should effectively scale up the incidence curve to better match
the targets.
\end{lstlisting}

Proposed parameters and structured residual feedback returned by
the harness:

\begin{lstlisting}[style=prompt]
proposed parameters (annual HSIL regression probabilities):
  y30_34 = 0.250  y35_39 = 0.250  y40_44 = 0.250  y45_49 = 0.250
  y50_54 = 0.250  y55_59 = 0.250  y60_64 = 0.250  y65_69 = 0.250
  y70_74 = 0.250  y75_79 = 0.250

error          : 2.822e-07
rmse           : 1.878e-04
bias           : 1.822e-04
max_abs_error  : 2.150e-04

  group     abs_error       bias            rel_bias_%
  y40_44    6.589e-05       6.589e-05         9.6
  y45_49    1.803e-04       1.803e-04        19.5
  y50_54    1.830e-04       1.830e-04        18.7
  y55_59    2.009e-04       2.009e-04        19.5
  y60_64    1.999e-04       1.999e-04        18.4
  y65_69    1.999e-04       1.999e-04        17.6
  y70_74    2.130e-04       2.130e-04        17.9
  y75_79    2.150e-04       2.150e-04        17.3
\end{lstlisting}

\subsection*{Iteration 2}

Reasoning block emitted by the agent:

\begin{lstlisting}[style=prompt]
The model currently overestimates cancer incidence across all
calibrated age groups, with relative biases ranging from ~10% to
~19%. Since increasing the HSIL regression probability reduces the
average time spent in the precancerous state and thereby lowers
cancer incidence, I will raise these parameters. To account for
the larger absolute errors in older strata and their higher
sensitivity to this parameter, I will apply a modest upward
gradient, increasing regression rates from 0.28 to 0.33 across the
age groups to better align the simulated curve with the targets.
\end{lstlisting}

Proposed parameters and structured residual feedback returned by
the harness:

\begin{lstlisting}[style=prompt]
proposed parameters (annual HSIL regression probabilities):
  y30_34 = 0.280  y35_39 = 0.280  y40_44 = 0.290  y45_49 = 0.300
  y50_54 = 0.300  y55_59 = 0.310  y60_64 = 0.310  y65_69 = 0.320
  y70_74 = 0.320  y75_79 = 0.330

error          : 1.707e-08
rmse           : 4.619e-05
bias           : -2.378e-05
max_abs_error  : 9.243e-05

  group     abs_error       bias            rel_bias_%
  y40_44    3.103e-06      -3.103e-06        -0.5
  y45_49    3.383e-05       3.383e-05         3.7
  y50_54    1.117e-05       1.117e-05         1.1
  y55_59    1.104e-06      -1.104e-06        -0.1
  y60_64    2.347e-05      -2.347e-05        -2.2
  y65_69    5.172e-05      -5.172e-05        -4.6
  y70_74    6.338e-05      -6.338e-05        -5.3
  y75_79    9.243e-05      -9.243e-05        -7.5
\end{lstlisting}

\subsection*{Final iteration}

Reasoning block emitted by the agent:

\begin{lstlisting}[style=prompt]
The largest remaining absolute errors are in y50_54
(underestimated) and y45_49 (overestimated), with smaller residual
biases in y55_59, y60_64, and y75_79. Since increasing the HSIL
regression probability reduces simulated incidence and decreasing
it raises incidence, I will slightly increase the rates for the
overestimated strata (y45_49, y55_59, y65_69, y70_74) and decrease
the rates for the underestimated strata (y50_54, y75_79, y60_64,
y40_44). These targeted micro-adjustments aim to balance the
signed biases and further minimize the RMSE.
\end{lstlisting}

Proposed parameters and structured residual feedback returned by
the harness:

\begin{lstlisting}[style=prompt]
proposed parameters (annual HSIL regression probabilities):
  y30_34 = 0.2800  y35_39 = 0.2800  y40_44 = 0.2874  y45_49 = 0.3210
  y50_54 = 0.2950  y55_59 = 0.3100  y60_64 = 0.2980  y65_69 = 0.3006
  y70_74 = 0.3020  y75_79 = 0.2985

error          : 4.914e-11
rmse           : 2.478e-06
bias           : -8.161e-08
max_abs_error  : 5.251e-06

  group     abs_error       bias            rel_bias_%
  y40_44    1.284e-07      -1.284e-07        -0.0
  y45_49    3.580e-06       3.580e-06         0.4
  y50_54    5.251e-06      -5.251e-06        -0.5
  y55_59    2.435e-06       2.435e-06         0.2
  y60_64    1.270e-06      -1.270e-06        -0.1
  y65_69    8.345e-07       8.345e-07         0.1
  y70_74    1.673e-07      -1.673e-07        -0.0
  y75_79    6.860e-07      -6.860e-07        -0.1
\end{lstlisting}

% =============================================================
\section{Representative iteration log: constrained calibration}
\label{app:log_excerpt_constrained}

The following excerpt is taken verbatim from a log produced during
the constrained calibration of the anal cancer model. The
constraint description was supplied to the agent through the
constraints section of the system prompt
(Appendix~\ref{app:constraint_prompt}); the description used in
this run is reproduced below. As in the unconstrained excerpt, the
reasoning blocks reproduced below were emitted by the agent at the
first two iterations and at the final iteration of the run, and
the residual tables are the structured feedback returned by the
harness. All proposals shown were accepted as feasible by the harness on
the first attempt.

\subsection*{Constraint description supplied to the agent}

\begin{lstlisting}[style=prompt]
HSIL regression probabilities should decrease or remain constant
with age.
\end{lstlisting}

\subsection*{Iteration 1}

Reasoning block emitted by the agent:

\begin{lstlisting}[style=promptctx]
The initial simulation consistently underestimates cancer
incidence across all age groups by approximately 23-25%,
indicating that the HSIL regression probabilities are too
high. Lowering these probabilities will increase the duration
individuals remain in the precancerous state, thereby raising
simulated incidence. I will reduce the baseline values and
introduce a slight decreasing trend with age to better
capture the rising target incidence curve,
(*@\textbf{while strictly maintaining the monotonicity constraint}@*).
\end{lstlisting}

Proposed parameters and structured residual feedback returned by
the harness:

\begin{lstlisting}[style=prompt]
proposed parameters (annual HSIL regression probabilities):
  y30_34 = 0.300  y35_39 = 0.290  y40_44 = 0.280  y45_49 = 0.270
  y50_54 = 0.260  y55_59 = 0.250  y60_64 = 0.240  y65_69 = 0.230
  y70_74 = 0.220  y75_79 = 0.210

error          : 4.569e-07
rmse           : 2.390e-04
bias           : 2.042e-04
max_abs_error  : 4.065e-04

  group     abs_error       bias            rel_bias_%
  y40_44    1.993e-06       1.993e-06         0.3
  y45_49    9.549e-05       9.549e-05        10.3
  y50_54    1.243e-04       1.243e-04        12.7
  y55_59    1.775e-04       1.775e-04        17.2
  y60_64    2.194e-04       2.194e-04        20.2
  y65_69    2.688e-04       2.688e-04        23.7
  y70_74    3.395e-04       3.395e-04        28.6
  y75_79    4.065e-04       4.065e-04        32.8
\end{lstlisting}

\subsection*{Iteration 2}

Reasoning block emitted by the agent:

\begin{lstlisting}[style=promptctx]
The simulation overestimates cancer incidence across all age
groups, with relative bias worsening sharply in older strata
(up to ~33% for y75_79). Since higher HSIL regression
probabilities reduce time in the precancerous state and lower
incidence, I need to increase these values, particularly for
older ages.
(*@\textbf{However, the constraint requires probabilities to be non-increasing with age}@*).
To resolve this, I will flatten the age-gradient
significantly, raising regression probabilities for older
groups (from ~0.21 to ~0.29) while keeping younger groups
relatively stable. This targeted increase should substantially
reduce the large errors in older ages with minimal disruption
to the already well-calibrated younger brackets.
\end{lstlisting}

Proposed parameters and structured residual feedback returned by
the harness:

\begin{lstlisting}[style=prompt]
proposed parameters (annual HSIL regression probabilities):
  y30_34 = 0.300  y35_39 = 0.299  y40_44 = 0.298  y45_49 = 0.297
  y50_54 = 0.296  y55_59 = 0.295  y60_64 = 0.294  y65_69 = 0.293
  y70_74 = 0.292  y75_79 = 0.291

error          : 5.986e-09
rmse           : 2.735e-05
bias           : 2.110e-05
max_abs_error  : 3.232e-05

  group     abs_error       bias            rel_bias_%
  y40_44    2.379e-05      -2.379e-05        -3.5
  y45_49    2.740e-05       2.740e-05         3.0
  y50_54    2.032e-05       2.032e-05         2.1
  y55_59    3.089e-05       3.089e-05         3.0
  y60_64    2.640e-05       2.640e-05         2.4
  y65_69    2.349e-05       2.349e-05         2.1
  y70_74    3.232e-05       3.232e-05         2.7
  y75_79    3.176e-05       3.176e-05         2.6
\end{lstlisting}

\subsection*{Final iteration}

Reasoning block emitted by the agent:

\begin{lstlisting}[style=promptctx]
The y40_44 age group continues to dominate the residual error
with a substantial negative bias (underestimation of cancer
incidence). To increase simulated incidence in this bracket,
the HSIL regression probability must be lowered.
(*@\textbf{However, the monotonicity constraint requires that regression probabilities do not increase with age}@*),
forcing any reduction in y40_44 to be matched by reductions
in adjacent older groups. I will lower
p_hsil_regression_annual for y40_44, y45_49, and y50_54 from
0.303 to 0.302, which aligns with the existing y55_59 value.
(*@\textbf{This maintains the strict non-increasing age trend}@*)
while pushing incidence upward in the youngest strata to
better match the target and reduce overall RMSE.
\end{lstlisting}

Proposed parameters and structured residual feedback returned by
the harness:

\begin{lstlisting}[style=prompt]
proposed parameters (annual HSIL regression probabilities):
  y30_34 = 0.307  y35_39 = 0.306  y40_44 = 0.302  y45_49 = 0.302
  y50_54 = 0.302  y55_59 = 0.302  y60_64 = 0.301  y65_69 = 0.300
  y70_74 = 0.299  y75_79 = 0.298

error          : 1.437e-09
rmse           : 1.340e-05
bias           : 1.128e-06
max_abs_error  : 3.229e-05

  group     abs_error       bias            rel_bias_%
  y40_44    3.229e-05      -3.229e-05        -4.7
  y45_49    1.428e-05       1.428e-05         1.5
  y50_54    3.684e-06       3.684e-06         0.4
  y55_59    9.997e-06       9.997e-06         1.0
  y60_64    3.046e-06       3.046e-06         0.3
  y65_69    1.043e-06      -1.043e-06        -0.1
  y70_74    6.500e-06       6.500e-06         0.5
  y75_79    4.856e-06       4.856e-06         0.4
\end{lstlisting}